\documentclass[sigconf]{acmart}
\setcopyright{none}
\usepackage{graphicx}
\usepackage{amsfonts}
\usepackage{amsthm}
\usepackage{algorithm}
\usepackage{multirow}
\usepackage{amsmath}
\usepackage{algpseudocode}
\usepackage{times}
\usepackage{latexsym}
\usepackage{booktabs}
\usepackage[T1]{fontenc}
\usepackage[utf8]{inputenc}

\usepackage{microtype}
\newcommand{\Tau}{\zeta}
\AtBeginDocument{%
  \providecommand\BibTeX{{%
    \normalfont B\kern-0.5em{\scshape i\kern-0.25em b}\kern-0.8em\TeX}}}

\acmConference[DRL4IR@CIKM2023]{Make sure to enter the correct
  conference title from your rights confirmation emai}{October 22, 2023}{Birmingham, UK}
\acmPrice{15.00}
\acmISBN{978-1-4503-XXXX-X/18/06}

\begin{document}

\author{Jianghong Zhou}

\affiliation{%
  \institution{Walmart Global Tech}
  \country{U.S.A.}
}
\email{jianghong.zhou@walmart.com}
\author{Joyce C. Ho}
\affiliation{%
  \institution{Emory University}
  \country{U.S.A.}
}
\email{joyce.c.ho@emory.edu}

\author{Chen Lin}
\affiliation{%
  \institution{Emory University}
  \country{U.S.A.}
}
\email{chen.lin@emory.edu}

\author{Eugene Agichtein}
\affiliation{%
  \institution{Emory University}
  \country{U.S.A.}
  }
  
\email{eugene.agichtein@emory.edu}

\title{ A Deep Reinforcement Learning Approach for Interactive Search with Sentence-level Feedback}
%% gap: U, offline-learning, cold-start
%% COSISTENCY OF NAME


%% 1. Limitation of assumptions
% Figure 6 SS?
% Table of hitl
% IBR?
% s -> F
% Table

\begin{abstract}
Interactive search can provide a better experience by incorporating interaction feedback from the users.
This can significantly improve search accuracy as it helps avoid irrelevant information and captures the users' search intents. 
Existing state-of-the-art (SOTA) systems use reinforcement learning (RL) models to incorporate the interactions but focus on item-level feedback, ignoring the fine-grained information found in sentence-level feedback.  
Yet such feedback  
requires extensive RL action space exploration and large amounts of annotated data.
This work addresses these challenges by proposing a new deep Q-learning (DQ) approach, DQrank. DQrank adapts BERT-based models, the SOTA in natural language processing, to select crucial sentences based on users' engagement and rank the items to obtain more satisfactory responses.
We also propose two mechanisms to better explore optimal actions.
DQrank further utilizes the experience replay mechanism in DQ to store the feedback sentences to obtain a better initial ranking performance.
We validate the effectiveness of DQRank on three search datasets. The results 
show that DQRank performs at least 12\% better than the previous SOTA RL approaches. We also conduct detailed ablation studies.
The ablation results demonstrate that each model component can efficiently extract and accumulate long-term engagement effects from the users' sentence-level feedback. This structure offers new technologies with promised performance to construct a search system with sentence-level interaction.
\end{abstract}
%%
%  Long term feedback - -query -- define, say it specifically  181~:
%  Try offline-online, (181-183)  State (good form) State = stored feedback
% define in abstract DQrank, definition of l
% Term-vageud-define  --> feedback form a query
\begin{CCSXML}
<ccs2012>
<concept>
<concept_id>10002951.10003317.10003338.10003343</concept_id>
<concept_desc>Information systems~Learning to rank</concept_desc>
<concept_significance>500</concept_significance>
</concept>
<concept>
<concept_id>10002951.10003317.10003347.10003354</concept_id>
<concept_desc>Information systems~Expert search</concept_desc>
<concept_significance>500</concept_significance>
</concept>
<concept>
<concept_id>10002951.10003317.10003347.10003348</concept_id>
<concept_desc>Information systems~Question answering</concept_desc>
<concept_significance>500</concept_significance>
</concept>
</ccs2012>
\end{CCSXML}

\ccsdesc[500]{Information systems~Learning to rank}
\ccsdesc[500]{Information systems~Expert search}
\ccsdesc[500]{Information systems~Question answering}
\keywords{reinforcement learning, neural networks, interactive search, sentence-level feedback}


\maketitle

\section{Introduction}
\begin{figure}
  \includegraphics[width=85mm]{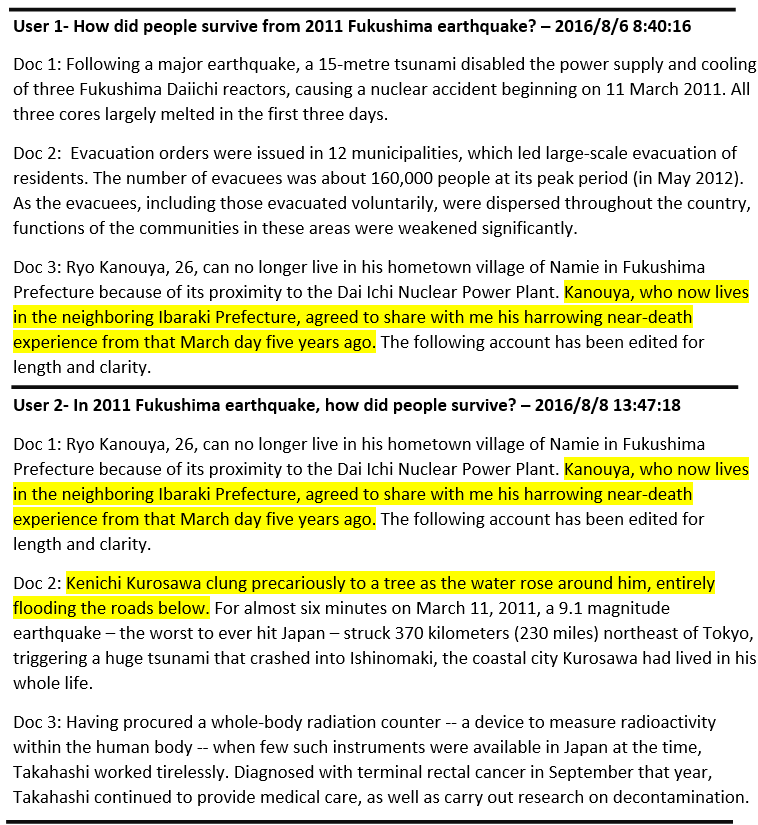}
  \caption{A use case of interactive search }
  \label{usecase1}
\end{figure}

Search with interactive settings can improve the performance of an information retrieval system \cite{ruotsalo2013supporting}. Users' interaction not only supplements the information of the query but also conveys users' latent intents. Users' feedback involves item-level feedback (e.g., item clicking, purchasing, and adding to cart) and sentence-level feedback (e.g., sentence clicking, copying, or reading). Sentence-level feedback can provide more details than item-level feedback. For example, in Figure \ref{usecase1}, User 1 searched 'How did people survive from the 2011 Fukushima earthquake' and selected (e.g., reading, clicking, or copying) a sentence from the $3$rd document as a piece of essential information. This interaction revealed that User 1 was looking for stories of survivors, helping User 2, who searched a similar query and obtained new results containing those stories based on User 1's feedback. When considering many people's interactions, the search results can approach the human's expected results. This scenario is common in question-answering systems (QA), e-commerce product reviews, and other complex search tasks \cite{liu2019study,zhou2020diversifying}. 

To build an interactive search environment, reinforcement learning (RL) approaches are the most common choices in the previous works \cite{zhao2019deep}. However, these studies suffer from two imperative challenges relating to (1) the amount of data needed to train the RL model and (2) the computational power needed to search the ample ranking space. We propose DQrank, a deep Q learning (DQ) based ranking system, to address these limitations.

First, obtaining data to train an interactive search model is strenuous due to the dynamic search environment. Directly depending on online users' interaction is costly and time-consuming, as users need to interact multiple times with the system to yield even semi-reasonable ranking results. On the other hand, solely relying on offline, static interaction records with only a limited set of all possible ranking results restricts the exploration of interactive search \cite{ruotsalo2018interactive}. Expanding the static interaction logs and combining these two learning ways are necessary. 
Previous research used strong assumptions to connect new ranking results to existing ones. \cite{weinzierl2020next} discounted or boosted the relevance between the query and document based on the position distance between the document and the clicked one. However, this approach can only relieve the effect of position bias but fail to reduce other bias, such as observation bias \cite{zhou2021biased}. \cite{ai2018unbiased} estimated the propensity to approximate the real relevance between the document and query, but still required encoding of the heuristic knowledge and can yield other biases. Furthermore, these methods focus on expanding the item-level connection, which cannot work on sentence-level feedback directly.
In this paper, we reduce the need for annotated data by adopting a BERT model \cite{liu2019roberta} pre-trained using tasks with explicit feedback to simulate user interactions. This user simulation function can then re-rank the items to obtain more satisfactory results. We also propose a novel state retrieval approach that combines both offline and online learning. By using similar queries as the starting point for RL, DQrank can obtain a better initial ranking performance. 
The second challenge, different from most RL tasks, lies in the notoriously gigantic action space of the ranking, so optimal exploration is infeasible. Previous research proposes different methods to approximate the optimal action that maximizes the target score. For example, RLIrank greedily ranks the documents with a neural-based agent \cite{zhou2020rlirank}. SlateQ decomposes the ranking to build their item choice model with mild assumptions \cite{ie2019slateq}. These methods can only obtain sub-optimal actions based on their choice mechanisms. Besides, they reformulate the ranking problems to point-wise problems without considering the effect of the surrounding items, which loses essential auxiliary information for the ranking evaluation and ignores widespread bias in real-life search tasks. To address the problem, we introduce a sliding window ranking method and rearrangement learning to explore the optimal actions. The sliding window prioritizes the top-ranking documents by focusing on a small range and pushing the significant items to the top position.
%\textcolor{red}{ranks the documents in a small range and prioritizes the top-ranking documents. It promises always to capture the optimal order in a small range and push the significant item to the top position.}
Additionally, rearrangement learning can adaptively learn the preferred order and continually improve the subsequent rankings. %\textcolor{red}{keep learning the preferred order and improve the following ranking.}

In summary, DQrank's contributions are three-fold:
\begin{itemize} 
    \item  DQrank contributes new methods, including self-supervised learning, BERT-based users' simulation, and state retrieval, to train the interactive search models, which expands the static search logs for the dynamic search environment and integrates offline and online learning.
       
    \item DQrank introduces novel sliding window ranking and rearrangement learning to efficiently explore the ranking results that can maximize the long-term value in DQ without losing the list-wise information of the ranking systems. 
    %REPHRASE: be concrete: DQrank proposes novel methods to improve the effectiveness of the RL-based search, including sliding window ranking and rearrangement learning method (REPHRASE),
    
    \item Additionally, we demonstrate our methods in open standard datasets and study the insights in a detailed ablation study.
    %REPHRASE: DQrank introduces self-supervised learning and data augmentations to train the Reinforcement Learning model without extensive online data. 

\end{itemize}

\section{Related Work}
\textbf{BERT-based ranking systems:} 
BERT is one of the most crucial text content embedding methods and thus is widely used in ranking systems \cite{vaswani2017attention}. Most BERT-based ranking systems are point-wise methods, which means the items are ranked by the similarity scores between the embedding results of the query and the document \cite{qiao2019understanding,nogueira2019passage}. Although these methods can utilize the query and document information, they do not learn the users' feedback from the interaction and ignore the ranking bias led by positions, observation, and other reasons \cite{gezici2021evaluation}. Those drawbacks may cause weaker ranking performance than list-wise and interaction-based ranking systems.

\noindent\textbf{Offline reinforcement learning to rank:} Offline RL methods use different RL frameworks to model the interactive search process. Because the models are normally trained by static search logs, these methods are mostly offline. In cooperation with premium embedding models, RL methods can outperform traditional learning-to-rank (LTR) models \cite{narayan2018ranking}. For example, RLIrank is a policy gradient method with a universal sentence encoder \cite{zhou2020rlirank,cer2018universal}. It outperforms important traditional LTR models like Adarank, ListNet and RankSVM \cite{xu2007adarank,xia2008listwise,chapelle2010efficient}.  Similarly, SlateQ, a Q learning-based method, is proposed to solve the ranking problem in recommendation systems. SlateQ achieves the best performance on RL-based recommendation approaches, also outperforming other non-RL approaches, including BERT-based models which demonstrates that DQ has the flexibility to learn a more complex state-action function without worrying about the unstability associated with training a nonlinear function approximator.
%Different RL methods are suitable for different tasks \cite{sutton2018reinforcement}. 
%For the interactive search, deep Q learning may be a better choice because the experience replay mechanism can explore static interaction records better \cite{sinha2022s4rl}. The proposed method is based on deep Q learning.

%The symbols we use in this paper is listed in Table \ref{tab:sym} in Appendix \ref{sec:appendix}. 

\section{Methodology}

This section proposes DQrank, a DQ model for sentence-based interactive search. In the interactive search setting, users search with a query and return sentence-level feedback like clicking or copying one or more sentences. 
In this section, we first introduce the DQ framework and how to apply it to interactive search. We then propose our sliding window ranking approach to explore the search space. 
Next, we present a method for identifying similar queries to avoid the cold-start problem.
Finally, we introduce self-supervised learning and data augmentation techniques to improve the generalizability and robustness of DQrank.
%Those interactions can be recorded or simulated. The simulation function can be trained by other interactive retrieval tasks, like question and answering or passage selecting tasks. 
%We also develop some essential methods to concrete our proposed RL framework.

%To make the discussion concrete, we focus on optimizing this interactive search process. 
%By rephrasing the query and the document sentences, we simulate the situation of different queries. The different users' situations can be resolved by collaborative filtering and Graph neural networks \cite{wang2019neural} \cite{su2009survey}, which are not a significant discussion of this topic. However, our model can still be applied to a typical situation like \ref{flowE} because we apply augmentation to generalize the training sets.

\subsection{Deep Q Learning Framework}

In RL, intelligent agents take actions in an environment to maximize the notion of cumulative reward \cite{kaelbling1996reinforcement}. In particular, DQ is an RL method that uses a deep neural network to approximate the action gain $Q$ \cite{fan2020theoretical}. Deep Q learning uses temporal difference (TD) learning to estimate the values $Q$ with previous actions sampled by an experience replay approach. The objective of temporal difference learning is to minimize the distance between the TD-Target and $Q(s, a)$, suggesting a convergence of $Q(s, a)$ toward its actual values in the given environment. 

In DQrank, we use DQ to model the interactive process for two reasons. First, DQrank has an experience replay mechanism, which is suitable for storing context information like feedback sentences and incorporating offline and online learning in interactive search. Second, the deep learning model in DQ can help better simulate the users' interaction, which can be used to expand the static search records. Therefore, DQ is appropriate for constructing our proposed model.

The DQ model consists of a set of states $S$, actions $A$, and a reward function $R$. In the context of the interactive search, the states $S$ reflect the \textit{query state}. This includes both query features and feedback sentences. We define the state $s$ as:
\begin{equation}
s_{t}=(q,f^{t}_1,f^{t}_2,...,f^{t}_E),
\end{equation}
where $s_{t}$ is the state at iteration $t$, $q$ is the query and $f_e^t$ is the $e$th feedback sentence at iteration $t$. $E$ is the number of (stored) feedback sentences. 
The query part will remain unchanged throughout the process, but the feedback part can be updated.
%\textcolor{red}{We define the State $s$ as:
%\begin{equation}
s%_{t}=(q,f^{t}_1,f^{t}_2,...,f^{t}_E)
%\end{equation}
%Where $s_{t}$ is the State at iteration $t$, $q$ is the query and $f_e^t$ is the $e$th feedback sentence at iteration $t$. $E$ is the number of stored feedback sentences.} 

The action space $A$ is the set of all possible ranking results. The initial retrieval will have some candidate items, $I$, and typically uses fast and straightforward methods to select these items from all possible items. 
%\textcolor{red}{After the initial retrieval, we assume we have some candidate items $I$. The initial retrieval typically uses fast and straightforward methods to select candidate items from millions of possible items.} 
Thus, the actions are the ordered subsets $A \subseteq I$ such that $|A|=N$, where $N$ is the number of items DQrank presents to the users.

To select the action and compute the Q value, we design a user simulation function $U$ for document ranking. The input of $U$ is a query $q$ and a sentence $f$ from the document. $U(q,f)$ measure the relevance of $q$ and $f$. While ranking, we calculate the $U$ of the first $M$ sentences of a document and select the highest score as the score for this document. 
%In this paper, we limit the $M\le50$. After that, we rank the documents based on this score. %The details of the $U$ are delineated in Section 3.3.

%Transition probability $P(s'|s,A)$ reflects the probability that State $s$ transitions to State $s'$ when action $A$ is taken.

Finally, the reward $R(s, a,s')$ is the reward of a list of ranking results, measuring user engagement or other metrics related to the ranking quality. $s'$ is the state before action $a$ and $a\in A$. %The details are discussed in Section 3.

Our task is to find a policy $\pi$ for the optimal ranking results. To achieve this, we organize the interactive search process as a DQ framework.
%In our deep Q learning, we want to maximize the total reward:
%$V^{\pi}=\sum^T_{t=0}{\gamma^t r_t}$, where $T$ is the total search and $r_t$ is the reward at the search $t$.
We use a BERT-based model $Q(s, a;\theta_i)$ to estimate the Q function in Q learning. $\theta_i$ can be obtained from the experience replay. 

The details of those parts are delineated in the following sections.

\subsection{Training}
\label{sec:offline}

We initialize a replay memory $Z$ for the experience replay in RL training and the feedback pool $P$ for the online serving system. All the transitions $(s_{t+1},a_t,r_t,s_{t})$ are stored in $Z$ and are randomly sampled as mini-batches to train the action-value function $Q$. The final states of each query are stored in the feedback pool $P$ after every episode. In each episode, we process one query. The feedback pool $P$ is used to accelerate and improve online learning.

The training of DQrank can be separated into two parts: (1) the offline training that can be done with a pre-trained users' simulation function, $U$, and (2) the online training to dynamically adjust to new search queries.
% Modify the sentence to:
% how to initialize the Q functions? Using the simulation function U or randomly

\textbf{Offline Training.}
The goal of the offline training is to learn the policy $\pi$ for the optimal ranking results based on the training dataset $W_q$. We first initialize the action-value function $Q$, the target action-value function $\hat{Q}$, and the state $s_1$.
%We then load a pre-trained users' simulation function $U$ and initialize the action-value function $Q$ and the target action-value function $\hat{Q}$. Those components will be discussed in the following sections.
%\textcolor{red}{When the training begins, we initialize the State $s_1$ and then}
Any fast and straightforward method can be used to conduct the initial ranking to obtain the initial ranking results $I_q$ for query $q$. 
%\textcolor{red}{The initial ranking uses a simple and fast approach to find the candidate items. The weak ranker we use in this paper is BM25 \cite{robertson2009probabilistic}.  }\textcolor{blue}{
For this paper, we use BM25 \cite{robertson2009probabilistic}. However, since the size of $I_q$ is still immense, we use $U$ to re-rank and obtain a smaller candidate set $T_q$. 
%After that, we load the training dataset $W_q$. 

At the beginning of an iteration, the agent selects an action $a$ based on the $\epsilon$-greedy policy $\pi$ :
  \begin{equation}
    \pi = \left\{
\begin{array}{rcl}
\pi_1 & & \text{with probability }\epsilon\\
\pi_2 & & \text{with probability } 1-\epsilon
\end{array} \right.
 \end{equation}
  $\epsilon$ is a parameter to balance exploration and exploitation \cite{rawson2021convergence}. 
 When the agent chooses $\pi_1$, we select a ranking result from $W_q$ if $W_q$ is not empty and then delete it from $W_q$. If $W_q$ is empty, we generate a random ranking result by selecting items from $T_q$. If the agent chooses $\pi_2$, we generate a ranking result from  $T_q$ and try to maximize the $Q$ value by re-ranking the results. In DQrank, we propose the sliding window ranking method see Section \ref{sec:window}, an efficient approach to explore the RL action space. %, which will be introduced in Section \ref{sec:window}.

After that, we can sample the transitions from $Z$ and obtain their data augmentations for the experience replay. The data augmentations can improve the robustness of the model and will be demonstrated in Section \ref{sec:da}. We minimize the difference between the TD-target and $Q$ value (Equation \ref{eq3}) in the experience replay and update the target function $\hat{Q}$ every $c$ step.
After finishing $T$ steps training, we store the best State of the query to the feedback pool $P$ for the online serving system. 

We then obtain the reward and store the transitions in $Z$. We later extract the samples from $Z$ for the experience replay. 

\textbf{Online Training.}
Online training aims to obtain the state $s$ for the new search queries. We first initiate the state $s_0$ with state retrieval, which can retrieve similar queries' states from the feedback pool $P$ and will be discussed in Section \ref{sec:state}. Then we go through the offline learning process without exploitation and model training. The feedback sentences $f$ in state $s$ are updated based on the users' feedback or interaction records. The transitions and final states are stored for the model refresh. Online training is more practical in industrious search tasks and much faster than offline training. 
% Why not directly use bert in the Q functions?
\subsection{The Users' Simulation Model}
\label{sec:user_sim}
The users' simulation function, $U$, mimics a user by determining whether the sentences in the document should be selected as the feedback sentence. In our model, $U$ is a BERT-based classification function. %\textcolor{red}{It stimulates the users and judges whether the sentences in the document should be selected as the feedback sentences.} 
Formally, the input of $U(q,f)$ is a pair of sentences. The first sentence $q$ can be the query or the selected feedback sentence. The second sentence, $f$, is a sentence from a document $D$.
Since the feedback sentences can reflect the users' satisfaction with the search results, the maximum probability
scores of all the sentences in the document can be considered a metric to rank the documents. 
%\begin{figure}
%\centering
  %\includegraphics[width=60mm]{deep_Q1.png}
  %\caption{The Structure of the users' simulation function $U(f,d)$. $f$ is the query or the selected feedback sentence. $d$ is the sentence from a document. $U$ represents the probability that $d$ is selected as the feedback for $f$. }
%  \label{fig2}
%\end{figure}
%The users' simulation function structure is presented in Figure \ref{fig2}. 
%\textcolor{red}{The input of $U(q,f)$ is a pair of sentences. The first sentence $q$ can be the query or the selected feedback sentence. The second sentence, $f$, is a sentence from a document $D$. When we are trying to access} 
Therefore, the relevance between the $q$ and a document $D$ can be modeled as $u(q,D) = \max_{f\in D} (U(q,f))$. 

For the input, $q$ and $f$ are separated by three tags, [CLS], [SEP], and [EOS], which represent the beginning, separator, and end of the sentence. The sentence embeddings are then encoded based on the tokens, segments, and positions as discussed in \cite{vaswani2017attention}. %\textcolor{red}{Then, they are encoded as embedding based on the tokens, segments, and positions. All the embedding techniques details can be found in \cite{vaswani2017attention}.} 
The embedding results from BERT, $x$, then represents the dependency between the $f$ and $d$. We feed $x$ to a fully connected neural network and calculate the selecting probability with the Softmax function, so $U = Softmax (W_T \times \tanh (x)+B_T)$. The  $U$ is further pre-trained by some question-answering and point-wise label datasets, which we introduce in Section \ref{sec:ss}.

%However, $U$ cannot reflect the overall performance of a ranking result because of the point-wise setting. In this setting, position bias, observation bias, and other factors are ignored, drastically affecting the users' satisfaction with the ranking results. Therefore it cannot be the $Q$ function we need for reinforcement learning. We further expand the model to a structure as Figure \ref{fig1}.

\begin{figure*}
  \includegraphics[width=\textwidth]{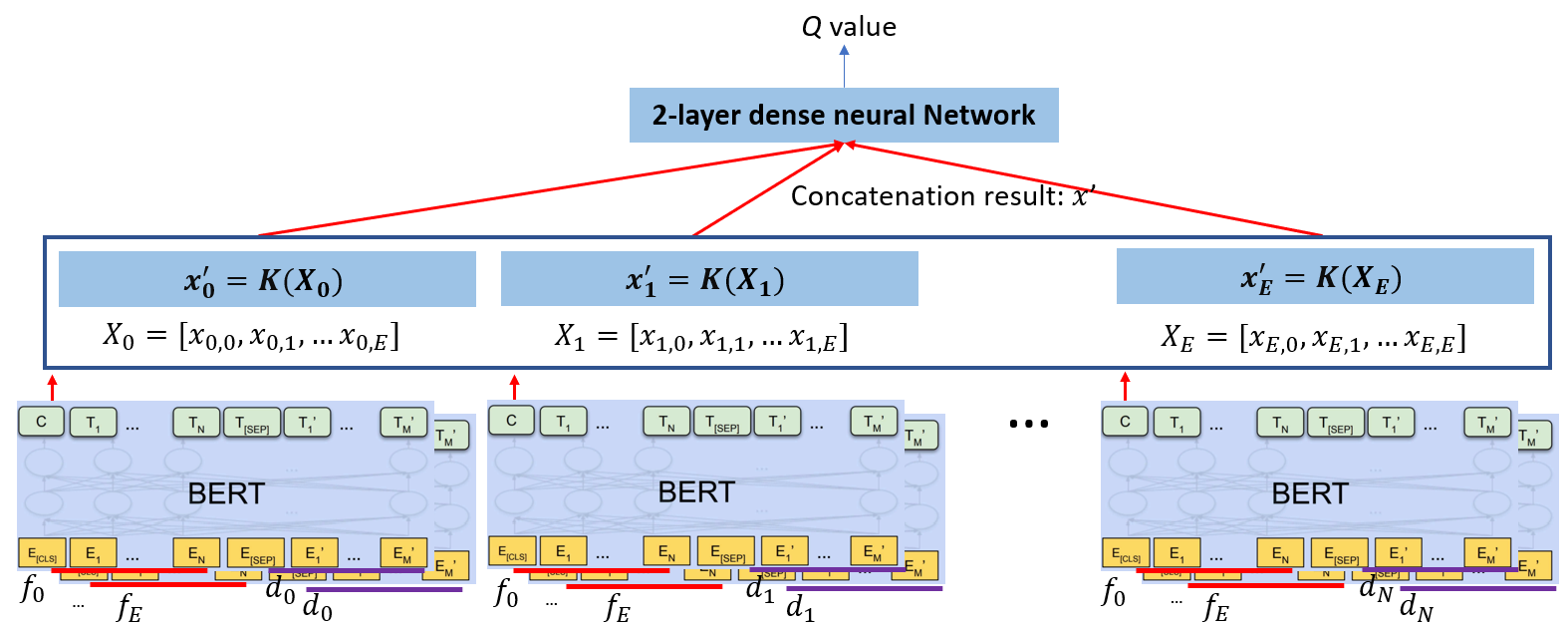}
  \caption{The Structure of Q function. $f_0$ is the query. $f_e$ is the $e$th feedback sentence. $E$ is the number of feedback sentences for the query $f_0$. $d_k$ is the sentence representing the $k$th document in the ranking results, such that $d_k = \text{argmax}_{d\in D_k} V(s,d)$, $s=(f_0,f_1,...,f_E)$. $k<=N$, $N$ is the number of documents in the ranking results.}
  \label{fig1}
\end{figure*}
%In this section, we adopt $SlateQ$ approach from the recommending systems to the ranking model \cite{ie2019reinforcement}. 

%{\bfseries Your document will be returned to you for revision if
%  modifications are discovered.}
\subsection{The Q Function}
The $Q$ function learns the value of the partial action (Figure \ref{fig1}), $a$ for a particular state, $s$. For ease, we denote $f_0$ as the query. The state is then $s=(f_0,f_1,f_2,...,f_E)$. The action $a$ is the list of ranked documents, $(D_1,D_2,...,D_N)$. For document $D_k$, we select the sentence $d_k= \text{argmax}_{d\in D_k} V(s,d)$ where $V$ is a discounted function based on the ranking metric as follows in Equation \eqref{equ:us}:
%\texcolor{red}{The structure of the $Q$ function is presented in Figure \ref{fig1}. The input of the function is the State $s$ and partial action $a$. The size of the partial action is decided by the sliding window size, which we discuss in Section \ref{sec:window}.  $s=(q,f_1,f_2,...,f_E)$. $f_0$ is the query. The Action $a$ is a list of documents  $a=(D_1,D_2,...,D_N)$. For the document $D_k$, we select the sentence $d_k= \text{argmax}_{d\in D_k} V(s,d)$. $V$ is a discounted function based on NDCG. NDCG is an important metric in ranking. It discounts the document's importance in different positions and evaluates the ranking results \cite{wang2013theoretical}. We define it in Equation \ref{equ:us}:}
%defined in Equation \ref{equ:us}. 

 %which i All the items in the ranking result are ordered by the discounted users' feedback $U_s(s_t, D)$ based on NDCG. NDCG is an important metric in ranking. It discounts the document's importance in different positions and evaluates the ranking results \cite{wang2013theoretical}.
 
% \begin{align}
% \begin{split}
%     &U_s(s_t,D)  = \max_{d\in D} u_s(s_t,d)
%     \end{split}
% \end{align}
 \begin{equation}
     V(s_t,d)=
     \frac{1}{\sum_{e=1}^{E+1}{\frac{1}{\ln(e+1)}}}\sum_{e=1}^{E+1}{[\frac{1}{\ln(e+1)}U(f_{e-1}^t,d)]},
     \label{equ:us}
 \end{equation}
where $d$ is a sentence of a document from the ranking result, $d_i^j$ represents the $j$th sentence of the document ranking at position $i$ in the searching results, and %$s_t=(f_0^t,f_1^t,f_2^t,...,f_E^t)$. $f_0^t=q$. 
$E$ is the number of feedback sentences.

Then we calculate the BERT embedding results,$x_{k,e}$,  for every pair of $f_e$ and $d_k$. Consequently, for every document $k$, we have $X_k = [x_{k,0},x_{k,1}...,x_{k,E}]$. We further weigh those embedding results:

\begin{align}
\begin{split}
    &x'_k = K(X_k) =
    \frac{1}{\sum_{e=1}^{E+1}{\frac{1}{\ln(e+1)}}}\sum_{e=1}^{E+1}{\frac{1}{\ln(e+1)}x_{k,e-1}}
\end{split}
\end{align}
 The weighted function $K$ is based on Normalized Discounted Cumulative Gain (NDCG), a popular approach to measure the quality of a list of ranking results \cite{jarvelin2002cumulated}. The proposed weighted function can re-balance the effect of the query and the feedback sentences when $E$ is of varying length. It also discounts the less important feedback sentences.  
 
 After the weighted embedding results are obtained, we concatenate them as the input to a 2-layer dense neural network used to estimate the $Q$ value. 
The weights of the $Q$ function can be obtained by minimizing the following loss function:
\begin{equation}
    L_i(\theta_i)=\mathbb{E}_{(s,a,r,s')\sim\rho(Z)}[(y_i-Q(s,a;\theta_i))^2],\label{eq3}
\end{equation}
where samples come from the replay memory $Z$ based on the random mini-batch sampling, $\rho$, and $ \theta_i$ are all the neural weights at iteration $i$. 

For every sample $(s_{j+1},a_j,r_j,s_{j})$, if $s_{j+1}$ is terminal, $y_j=r_j$, otherwise 
\begin{equation}
    y_j = r_j+\gamma \max_{a'}\hat{Q}(s_{j+1},a';\theta^-),
\end{equation}
 and weights of target action-value function $\hat{Q}$, $\theta^-$ is updated to $\theta$ every $c$ steps. Since the BERT model is optimized multiple times in one step, we calculate the average values as the weights for the next step.

\subsection{The Reward Function}

The reward $r$ evaluates the searching result $a$ given the query $q$. Therefore, $r = R(s,a,s')=R(a,q)$. For some search results, we can find them in the dataset, $W_q$, and are already labeled. However, the agent generates many search results, and not all the labels exist in the dataset. To calculate the reward, we propose a reward transition method.

To estimate the reward $\hat{r}$ of an action $a$, first we extract a existed ranking result $a'_i$ from the dataset $W_q$. We can calculate its DCG score with its query-document relevance scores. If the scores are not provided, we can use the user simulation function $U(q, D), D\in a'_i$. We assume the reward is a position-discounted score like NDCG, then we have
\begin{align}
\begin{split}
&\hat{R}(a,q) = \xi_i R(a'_i,q),\\
    &\xi_i=\frac{\hat{R}(a,q)}{R(a'_i,q)}\approx\frac{\text{NDCG}(a,q)}{\text{NDCG}(a'_i,q)}=\frac{\text{DCG}(a,q)}{\text{DCG}(a'_i,q)}\\
    &=\frac{\sum_{D_k\in a}\frac{u(q,D_k)}{\ln(k+1)}}{\sum_{D_k\in a'_i}\frac{u(q,D_k)}{\ln(k+1)}}
\end{split}
\end{align}

 We can use the average estimation over $W_q$ as our reward:
\begin{equation}
    \hat{R}(a,q) = \mathbb{E}_{a'_i\sim W_q}(\xi_i R(a'_i,q)),
\end{equation}
where $W_q$ is the ranking dataset without deleting any element.

\subsection{Sliding Window Ranking}
\label{sec:window}
 Selecting an action to maximize the $Q$ value is notoriously difficult for the ranking tasks \cite{ie2019reinforcement}. To maximize the Q-value, we have to check every combination of the candidate documents. This colossal search space cannot be deployed to meet the latency requirements in most ranking systems. To effectively explore the searching space, we develop the sliding window approach by considering both users' simulation $U$ and the $Q$ functions.
 As introduced in Section \ref{sec:user_sim}, $U$ can reflect the users' satisfaction with the item but cannot demonstrate the overall ranking performance. However, it is still helpful to generate approximate ranking results. Therefore, the first step is ranking the candidate documents with $U$.
 
\begin{figure}
  \includegraphics[width=\linewidth]{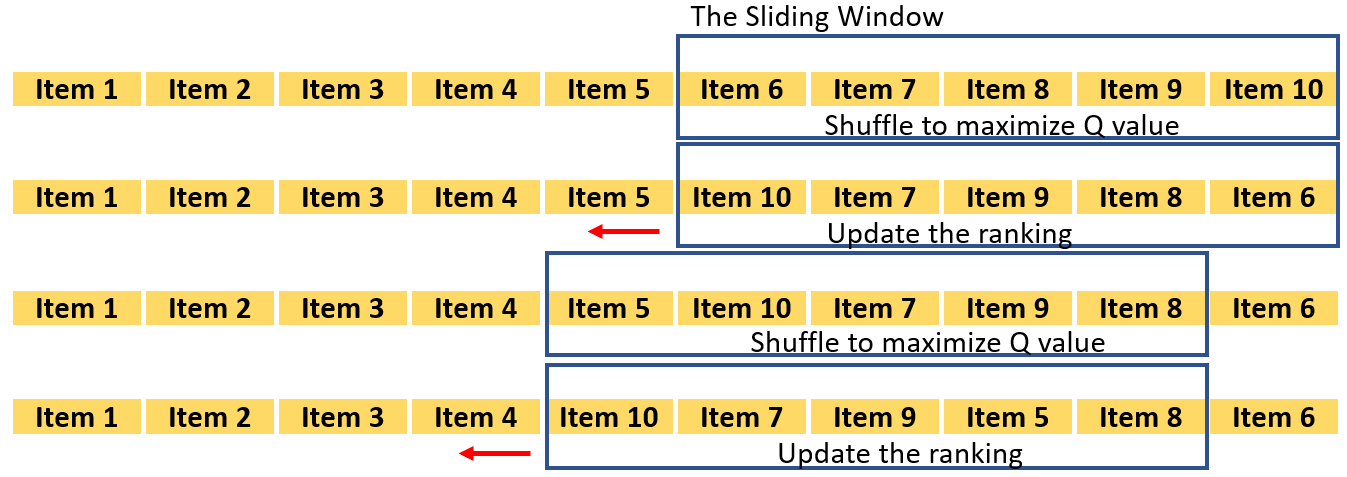}
  \caption{The Sliding Window Ranking Workflow }
  \label{fig3}
\end{figure}

Given the users' ranking, we design a sliding window of $m$, a parameter that reflects the users' attention. %\textcolor{red}{the size that can accommodate the presenting items.} 
We start the window at the last item and move it backward, item by item, until we encounter the first item. Within each sliding window, we evaluate every combination to obtain the maximal Q value within the window. We replace the original order with this optimal order and proceed to the next window. The workflow of the sliding window ranking is presented in Figure \ref{fig3}.

  The advantages of this proposed method are three-fold:
 \begin{enumerate}
    \item This approach models the attention of the users. Users' attention is typically limited to a few items, so the position and observation bias is commonly generated within an attention window. Since we rank the items with the users' simulation function $U$ first, it is reasonable to reflect the effect of ignored biases.
    
    \item In most ranking tasks, the top-ranking items are usually the most important because users only browse the first serving page. This approach prioritizes the top-ranking items by ensuring items mistakenly ranked in the back can be moved forward.
    %\textcolor{red}{by moving the window backward. Those items mistakenly placed in the back can be moved ahead step by step.}
    \item Moreover, this approach is efficient enough to be deployed in the actual training process. If we assume that we have $G$ items to rank and the number of our presenting items is $N$, we, in theory, would need to evaluate $C(G,m)$, where $C$ denotes the number of combinations. However, under our approach, we only need to try it $(G-m+1)m$ times, which is $\frac{1}{G-m+1}C(G,m)$ times faster.
    %\textcolor{red}{we need to try $P(G,m)$ times. $P$ means the permutation. With this approach, we only need to try $(G-m+1)P(m,m)$ times, which is $\frac{1}{G-m+1}C(G,m)$ times faster. $C$ means the combination.} 
    Since $m$ is usually relatively small, the computation time of this ranking method can be controlled in a reasonable scope.
\end{enumerate}
 
 To accelerate the computation, we calculate the embedding results and then apply this sliding window ranking method in the concatenation part to obtain different $x'$. By comparing the $Q$ value, we can find the approximate optimal action.
 
\subsection{Rearrangement Learning }
\label{sec:re}

Since users' feedback is mainly implicit, as the ranking results are usually passively digested \cite{agarwal2019general}, there can be discrepancies between the ranking using the users' simulation function, $a_U$, and the ranking results that maximize the $Q$ value, because some relevant documents ignored by the users can be annotated as irrelevant. This inconsistency can introduce noise to the training and cause slow convergence and poor performance. 
%\textcolor{red}{The users' simulation function $U$ can help us simulate users' selecting behaviors. However, users usually passively digest the ranking results in the ranking tasks. That feedback is largely implicit \cite{agarwal2019general}. Therefore, the ranking results based on the simulation $a_U$ can be very different from the ranking results $a_Q$ that maximize the $Q$ value.} 
Since we use $U$ to estimate the optimal action and rank the documents, it is crucial to make $a_U$ conform to $a_Q$. To achieve this, we train the function $U$ by minimizing the loss function:
\begin{align}
L(\omega) =
\mathbb{E}_{d_{Q,j}\sim a_Q,d_{U,j}\sim a_U,f\sim S}[(U(f,d_{Q,j};\omega)-y_{U,j})^2], \label{eq:u_l}
\end{align}
%\begin{align}
%\begin{split}
%    &L(\omega) = \\
%    &\mathbb{E}_{d_{Q,j}\sim a_Q,d_{U,j}\sim a_U,f\sim S}[(U(f,d_{Q,j};\omega)-y_{U,j})^2]\label{arl}
%\end{split}
%\end{align}
where $d_{Q,j},d_{U,j}$ are the the sentences selected to represent documents $D_{Q,j},D_{U,j}$ respectively; $D_{Q,j},D_{U,j}$ are the $j$th documents in the ranking $a_Q$ and $a_U$ respectively; $f$ is the query or the feedback sentences from state $s$, and $y_{U,j}= U(f,d_{U,j})$.

The rearrangement of $U$ proposed in Equation \eqref{eq:u_l} can be seen as a self-inference method to estimate the difference between simulated score $U$ and the optimal score $\hat{U}$ that can rank the items to maximize the $Q$ value. $U$ is an auxiliary $Q$ function to help us rank the items because directly using the $Q$ function to rank the document is computationally impossible. Since $U$ cannot incorporate some of the features from the other items, it cannot accurately estimate the $Q$ value. However, rearrangement learning helps us model a probability distribution $\delta = \hat{U}-U$ that can close the gap between  $U$ and $\hat{U}$. Noticeably, $\mathbb{E}(\delta)=0$, because
$\mathbb{E}_{d_{Q,j}\sim a_Q,d_{U,j}\sim a_U,f\sim S}[U(f,d_{Q,j})- U(f,d_{U,j})]$
$=0.$
%\begin{align}
%\begin{split}
%&\mathbb{E}_{d_{Q,j}\sim a_Q,d_{U,j}\sim a_U,f\sim S}[U(f,d_{Q,j})- U(f,d_{U,j})]\label{arl}\\
%&=0
%\end{split}
%\end{align}

\subsection{State Retrieval}
\label{sec:state}
The current framework assumes that users provide feedback throughout the entire reinforcement process, which in our experiments requires at least 15 iterations to converge. A user's search can be considered the first iteration of the RL with $\epsilon=0$.
In real-life search tasks, it is hard for the users to interact with the search engine for more than a handful of iterations. Thus, we propose to use state retrieval to identify similar queries to serve as the starting point for $s_0$.
%\textcolor{red}{go through the whole reinforcement process. Users' search can be considered the first iteration of the RL with $\epsilon=0$. The return result is the action $a$ maximize $Q$ based on sliding window ranking.}

Formally, let us denote the final state for query $q$ as $s_T$. We store the BERT-encoded search query, $x(q)$, and $s_T$ in our feedback pool $P$.
When a new query is provided, $q'$, we use BERT to encode the query, $x(q')$, and calculate the cosine similarity between this and all queries stored in $P$.
%\textcolor{red}{
%However, we can also consider that the search begins from the final State. We store those states in $P$. In actual search tasks, users searching for similar or similar queries are common. We use the distilBERT model to encode the search query and calculate the cosine similarity between the query stored in $P$.} 
If the highest cosine similarity is higher than our setting threshold $\psi$, we can retrieve this state as the initial state of the query. Consequently, the feedback sentences generated before can help improve the search.

%\textcolor{red}{Local exploration with data augmentations helps our model become more robust and stable. Even though the queries may not be semantically identical, the relevant feedback sentence can still provide extra knowledge to rank the documents. The experiments in Section \ref{sec:45} show that this retrieval helps improve performance.}

\subsection{Self-supervised Learning}
\label{sec:ss}
% Data augmentation is a common method to train the neural network in computer vision research \cite{wei2019eda}. Recent research also expands it to offline reinforcement learning, mostly robotic control tasks \cite{sinha2022s4rl}. Those research support that data augmentation can help local exploration of the trajectories in the dataset. 
Providing sentence feedback is essential in our users' simulation and relevance analysis. However, it is costly and challenging to collect sufficient annotations where sentences are labeled that help the search.  Therefore, we use self-supervised learning to solve this issue.
 
 Self-supervised learning is a method that can reduce the data labeling cost and leverage the unlabelled data pool \cite{hendrycks2019using}. In this paper, we pre-train the users' simulation function $U$ with the data of Question-answering (QA) systems. QA systems are sentence-based search systems. Only a few sentences in a document are considered the critical answer to the question. In this paper, we train the model by estimating the importance of the sentences in the document. Those sentences are labeled as 'Selected' or 'Not Selected.'
 After being pre-trained by the QA systems, the function $U$ can be used to label the sentences crucial to the search, which are essential to training the DQ model further.
 
\subsection{Data Augmentation}
 \label{sec:da}
 During the training process, DQrank is only exposed to a static set of queries (i.e., $W_q$). However, in practical settings, the search terms can be slightly different yet should still yield similar results. Therefore, to enhance the model's ability of generalization and robustness, we propose local exploration with data augmentations. This is done by reorganizing the search process of different sessions as an offline RL process to utilize the long-term benefit of the feedback sentences.
 %\textcolor{red}{In DQrank, we reorganize the search process of different sessions as an offline RL process to better take advantage of the long-term benefit of the feedback sentences. In a training iteration, the query is static in different steps. However, in the actual search tasks, if the search sessions are conducted at a different time or with different users, the search terms can be slightly different. Therefore, to keep DQrank practical, we need to enhance the model's ability of generalization and robustness. Local exploration with data augmentations is of vital importance.}   
 
We design a new transformation $\hat{S_t}=\Tau(s_t)$, where $\hat{S_t}$ is the data augmentations of $s_t$, and the reward is smooth, that is $r(s_t)\approx r(\hat{s_t})$. Note that the data augmentation cannot be too aggressive. Otherwise, it may hurt the performance of the model, because the reward for the original state does not coincide with the reward of the augmented state (i.e., $r(s_t,a_t)\neq r(\hat{s_t},a_t),\hat{s_t}\in\hat{S_t}$). Therefore, the choice of $\Tau$ needs to consider the connection between the reward and the state. 
 
 Since we cannot change the semantics of the original text and need to maintain the reward function remains smooth, we choose to paraphrase the sentences. ``Paraphrase" expresses the meaning of a sentence using different words.  %An example of Parrot is presented in Figure \ref{fig4}.
 %In recent years, many premium and commercial transformer-based paraphrasing methods have developed, such as  QuillBot\cite{fitria2021quillbot} and Rasa \cite{bocklisch2017rasa}. Some experimental and advanced tools, like NLPAug \cite{ma2019nlpaug} and NL-Augmenter \cite{dhole2021nlaugmenter}, also achieve some great performances. After considering three crucial aspects of paraphrasing, which are adequacy (Is the meaning preserved adequately?), fluency (Is the paraphrase fluent in English?), and Diversity (How much has the paraphrase changed the original sentence?), 
 %\begin{figure}
%  \includegraphics[width=77.5 mm]{Para.png}
%  \caption{An example of the paraphrase sentences generated by Parrot. }
%  \label{fig4}
%\end{figure}
  We use the paraphrase technique to generate many queries with similar meanings to the query that goes through the RL framework. Those queries are used in the pre-training process of the users' simulation function $U$ and the experience replay. We choose an advanced paraphrase toolbox Parrot as our paraphrasing model \cite{prithivida2021parrot}.
 
 To reduce some noises and stabilize the training, we average the state-action values and target values over different data augmentations of the state. We let
 
 \begin{equation}
 Q_t(s_t,a_t) = \frac{1}{|\hat{S_t}|+1}\sum_{i=0}^{|\hat{S_t}|}Q( \Tau_i(s_t),a_t)),
\end{equation}
where $|\hat{S_t}|$ is the number of the data augmentations, $\Tau_0(s_t)=s_t$, and $\Tau_i(s_t),i\neq 0$ is the data augmentation of $s_t$. Since we assume the reward is locally smooth, $r_t$ in Equation \eqref{eq3} remains the same.

%\begin{figure*}
%  \includegraphics[width=\textwidth]{deep_Q.png}
%  \caption{The Structure of Q function}
%  \Description{All the BERTs are the same with simulation function $U$. $f_0$ is the query. $f_e$ is the $e-1$th feedback sentence. $E-1$ is the number of feedback sentences for the query $f_0$. $d_k=\argmax_s_{k} U(f_0,s_k)$, $s_k$ are the sentence from $k$th document. $k<=N$, $N$ is the number of documents in the ranking results.}
%  \label{fig3}
%\end{figure*}
\begin{table*}
\centering
  \caption{The comparisons on the ranking performance. The best results are in bold. Results marked with * indicate significant improvements with p<0.05 than others.}
  \label{tab:commands}
  \begin{tabular}{lcccccc} 
  \toprule[2 pt]
  \multicolumn{1}{c}{\multirow{2}{*}{Model}}&\multicolumn{2}{c}{MS-MARCO}&\multicolumn{2}{c}{ORCAS}&\multicolumn{2}{c}{HITL}\\ 
   \cmidrule(l r){2-3}\cmidrule(l r){4-5}\cmidrule(l r){6-7}
     &nDCG@10&MRR&nDCG@10&MRR&nDCG@10&MRR\\
    \midrule[1.5 pt]
      BM25&0.5246&0.2704&0.5726&0.2844&0.3424 &0.2133\\ BERT-U&0.5382&0.2871&0.5897&0.2974&0.3623 &0.2245\\     
      \midrule
      RLIrank&0.5627&0.3004&0.6146&0.3116&0.3829 &0.2363\\ 
      SlateQ&0.5573&0.2987&0.6068&0.3104&0.3971 &0.2411\\ 

     \midrule
     DQrank (BERT)&0.6221&0.3145&0.6424&0.3287&0.4422 &0.2754\\ 
     
     DQrank (DistilBERT)&0.6154&0.3127&0.6336&0.3265&0.4402 &0.2733\\ 
     DQrank&\bf{0.6313*}&\bf{0.3224*}&\textbf{0.6517*}&\textbf{0.3349*}&\textbf{0.4458*} &\bf{0.2773*}\\ 
    \bottomrule[2 pt]
  \end{tabular}
\end{table*}
\section{The Experimental Setting and Results}

In this section, we design experiments to demonstrate that DQrank can achieve better ranking performance in search tasks.
\subsection{Datasets}
 We evaluate our model's ranking performance with three datasets. Those datasets evaluate the ranking results in different ways. The details are delineated as follows:

\noindent\textbf{MS-MARCO}: MS-MARCO datasets are large datasets designed for different information retrieval tasks \cite{nguyen2016ms}. We select the document retrieval dataset to evaluate the proposed model. 
This dataset contains 3.2 million documents and 
 provides relevance scores for every pair of queries and documents. Therefore, we can calculate the overall performance of a ranking result with the normalized discounted cumulative gain (NDCG) score. At the same time, we use the passage retrieval dataset to fine-tune the users' simulation function $U$. The passage retrieval dataset has 8.8 million passages.

\noindent\textbf{ORCAS}: ORCAS is a click-log based dataset \cite{craswell2020orcas}.  This dataset uses Indri \cite{strohman2005indri} to retrieve 100 documents for every search. The clicks are used to evaluate these ranking results. The collection contains 20 million clicks, 1.4 million documents, and 10 million queries.
    
\noindent\textbf{HITL\footnote{\url{anonymous.com}}}: The human-in-the-loop (HITL) dataset is an interactive search dataset from a CLIR (Cross-linguistics Information Retrieval) project. In this dataset, there are eight topics and eight queries per topic. For each query, volunteers are asked to select some relevant sentences from the top 10 documents as feedback. The number of documents is 30000, and 1053 sentences are selected as feedback sentences. The dataset belongs to a multi-university collaborative project in the USA and will be released soon.

We also use two additional datasets to fine-tune the users' simulation function $U$: ASNQ and ciQA. ASNQ is a high-quality natural language dataset for answer sentence selection, which consists of 59914 questions  \cite{garg2019tanda}. ciQA is a web-based dataset for training interactive question answering \cite{kelly2007overview}. It contains 30 topics.
    
\subsection{Experimental Settings}
 We use mean reciprocal rank (MRR) and NDCG of the top 10 items (nDCG@10) to measure the ranking performance. We use two point-wise ranking approaches and two state-of-the-art (SOTA) RL methods in search or recommendation systems as the baselines. Additionally, we evaluate the performance of three versions of our proposed model with different BERT models. The baselines include:
 \begin{itemize}
     \item \textbf{BM25} \cite{robertson2009probabilistic}: Baselines of the experiments .
     \item \textbf{BERT-U}: The proposed BERT-based user simulation model ranks the documents directly.
     \item \textbf{RLIrank} \cite{zhou2020rlirank}: RLIrank is a policy gradient-based RL method, which is SOTA RL-based interactive search models.
     \item \textbf{SlateQ} \cite{ie2019slateq}: SlateQ is a Q learning method with decomposition settings, which is SOTA RL-based recommendation systems.
     \item \textbf{DQrank (BERT)}: The DQrank model with $U$ pre-trained using BERT \cite{vaswani2017attention}.
     \item \textbf{DQRank (DistilBERT)}: The DQrank model with $U$ pre-trained using DistilBERT, a smaller transformer version trained by distilling BERT \cite{sanh2019distilbert}.
    \item \textbf{DQrank}: The DQrank model with $U$ pre-trained using RoBERTa, a BERT model trained with a larger dataset \cite{liu2019roberta}. It is also our proposed implementation of DQrank.
 \end{itemize}
  The input embedding and structure settings are based on the \cite{wolf-etal-2020-transformers}.
  
  We use 5-fold validation to examine our approach \cite{jung2018multiple}. The data augmentations are only applied to the training part.
    The hyperparameters are determined by grid search \cite{marutho2018determination}. We use the elbow method to determine the sliding window size, $m=4$. We use the Adam optimizer to train the model and the learning rate of 0.001.

\subsection{Results}

The ranking performance of the various models is presented in Table \ref{tab:commands}. We can observe that DQrank has more than 12\% improvements over the previous SOTA RL approaches. Additionally, we found that the basic BERT-U model only has limited improvement from the benchmark BM25 ($+2.6\%$), while the DQ framework brings more significant gains in ranking performance ($+15.6\%$ when comparing DQrank (BERT) with BERT-U). This observation indicates that the proposed learning framework and feedback sentence information are crucial for a better search system. 

In terms of RLIrank and SlateQ, they outperform BM25 and BERT-U. However, they only have limited improvement ($7.26\%$), demonstrating that the policy gradient approach and the DQ method without DQrank settings cannot utilize the interaction records sufficiently enough to drastically improve the interactive ranking performance. 

Additionally, we compare the performance of the DQrank with different transformers. We can find that DQrank (our proposed implementation with RoBERTa) achieves the best performance. However, it only yields a $2.5\%$ improvement over the other transformers. These results shows that a relatively lightweight BERT model, like distilBERT, can potentially replace RoBERTa without hurting the ranking performance. This can further accelerate DQrank and make it more practical and efficient in industrial search tasks.

\subsection{Ablation Study \label{sec:45}}
In this subsection, we investigate the effects of four essential components of DQrank: data augmentations (DA), state retrieval (SR), rearrangement learning (ARL), and self-supervised learning (SS). The models are listed as follows:
\begin{itemize}
    \item \textbf{DQrank}: The RoBERTa-based DQrank model.
    \item \textbf{DQrank-SR}: The DQrank without the state retrieval mechanism (SR) introduced in Section \ref{sec:state}.
    \item \textbf{DQrank-DA}: The DQrank without data augmentation training (DA) proposed in Section \ref{sec:da}.
    \item \textbf{DQrank-DA-SR}:  The DQrank without both SR and DA.
    \item \textbf{DQrank-SS}: The DQrank without self-supervised learning (SS) discussed in Section \ref{sec:ss}.
    \item \textbf{DQrank-SS-DA}: The DQrank without both SS and DA.
\end{itemize}
The results with and without rearrangement learning (see Section \ref{sec:re}) are presented in Table \ref{tab:commands2}. With rearrangement learning, DQrank-SR, which only uses data augmentations, provides an increase of 5.21\%  in nDCG@10 and 2.65\% in MRR. When only using state retrieval (DQrank-DA), we increase 2.73\% in nNDCG@10 and 1.32\% in MRR. 

We found that both state retrieval and data augmentation can improve search performance, and the latter is more helpful. However, when incorporating both approaches, we increase 9.58\% in nDCG@10 and 4.64\% in MRR, which is a significant improvement.
This analysis reveals that data augmentations and state retrieval have substantial joint effects. The state retrieval can help us re-use those proper feedback sentences, and the local exploration from the data augmentations can connect those sentences to relevant documents.

We also observe that all the scores decrease an average of 8\% when we train the models without rearrangement learning. This observation shows that it is necessary to align the user simulation function ranking to maximize the $Q$ value, as the gap can cause discrepancies to arise between the ranking results and yield suboptimal ranking results. Thus,
rearrangement learning is significant for the users' simulation function $U$ to support the documents' ranking.
\begin{table*}
\centering
  \caption{The ranking performance of DQrank with different components in the MS-MARCO dataset.     ARL is rearrangement learning. All the results have significant improvements with p<0.05 than DQrank-DA-SR without ARL.}
  \label{tab:commands2}
  \begin{tabular}{lcccccccc} 
  \toprule[2 pt]
  \multicolumn{1}{c}{\multirow{2}{*}{Model}}&\multicolumn{2}{c}{Without ARL}&\multicolumn{2}{c}{With ARL}\\ 
   \cmidrule(l r){2-3}\cmidrule(l r){4-5}
     &nDCG@10&MRR&nDCG@10&MRR\\
    \midrule[1.5 pt]
      DQrank-DA-SR&0.5406&0.2745&0.5677 (+5.01\%)&0.2977 (+8.45\%)\\ 
      \midrule
      DQrank-DA&0.5554 (+2.73\%)&0.2822(+2.80\%)&0.5832 (+7.88\%)&0.3017 (+9.90\%)\\ 
       \midrule
    DQrank-SR&0.5567 (+2.98\%)&0.2831 (+3.13\%)&0.5973(+10.49\%)&0.3056 (+11.32\%)\\ 
     \midrule
    DQrank&0.5597 (+3.53\%)&0.2865 (+4.37\%)&0.6221 (+15.08\%)&0.3145 (+14.57\%)\\
    \bottomrule[2 pt]
  \end{tabular}
\end{table*}

\begin{figure}
  \includegraphics[width=\linewidth]{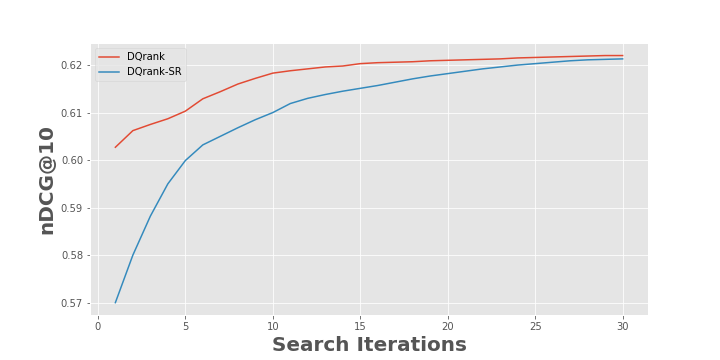}
  \includegraphics[width=\linewidth]{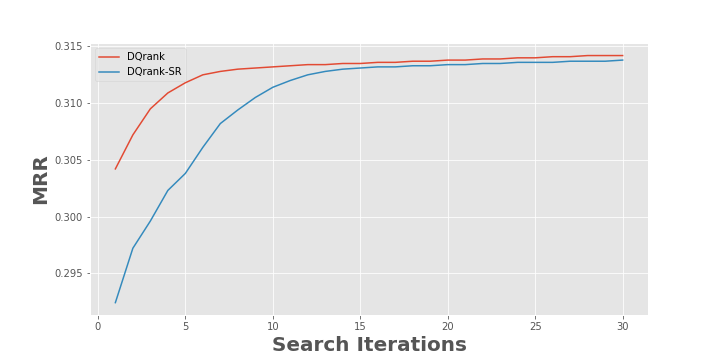}
  \caption{The ranking performance of DQrank with (red) or without state retrieval (blue) during the search session in the MS-MARCO dataset.  }
  \label{bar1}
\end{figure}

Additionally, we explore the effect of state retrieval on the ranking performance during the online search sessions. As shown in Figure \ref{bar1}, in the initial search, both nDCG and MRR increase quickly from 0 to 5 iterations. However, DQrank has higher initial scores than DQrank-SR, which are +5.7\% in nDCG@10 and +5.2\% in MRR. After 30 search sessions, both models converge to relatively stable states, but DQrank is still around 1\% better than DQrank-SR. This ablation study reveals that 
%\textcolor{red}{state retrieval can help the online DQrank have a better initial ranking performance. } 
the feedback sentences collected from the offline search logs can help DQrank obtain better final states.

\begin{figure}
  \includegraphics[width=\linewidth]{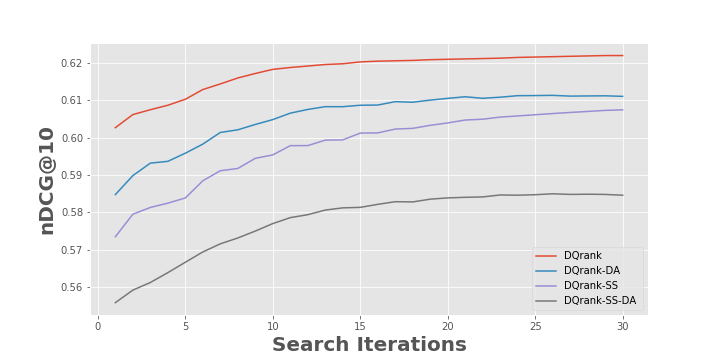}
  \includegraphics[width=\linewidth]{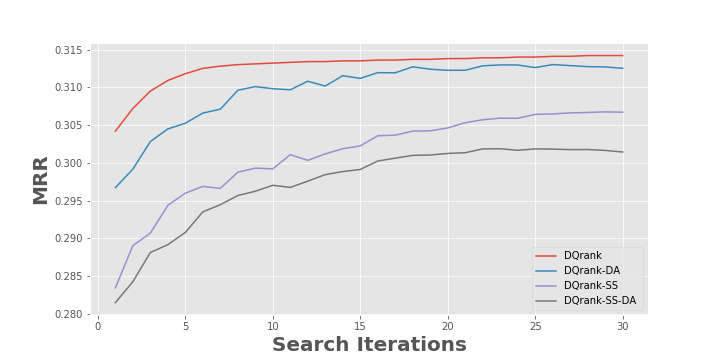}
  \caption{The ranking performance of DQrank with different components, DQrank (red), DQrank-DA (blue), DQrank-SS (purple), and DQrank-SS-DA (black), in the MS-MARCO dataset.}
  \label{pre}
\end{figure}

Lastly, we study the effect of self-supervised learning and data augmentations as a function of the number of search iterations $t$. The results are presented in Figure  \ref{pre}. We observe that data augmentations and self-supervised pre-training can improve the initial and final ranking performance by an average of 4\%. At the same time, they have a combined effect of increasing both NDCG@10 and MRR by 6\%. This ablation study shows that data augmentation and self-supervised learning can effectively boost the initial search performance while supporting the improved final search performance. 
\section{Conclusion}
This paper proposes a new RL approach for the search systems with sentence-level feedback, DQrank. This method aggregates similar search sessions as an interactive process and optimizes the search results with DQ by considering the 
feedback sentences. Additionally, we provide three significant contributions to improving this RL framework: (1) We propose sliding window ranking to estimate the optimal action efficiently. (2) We introduce state retrieval to re-use the feedback sentences from the search history. (3) We reformulate the query using data augmentation to help the RL agent explore locally, making the search process stable and robust.
%\textcolor{red}{
%(1) We first introduce query reformulation as data augmentation to help the RL agent explore locally, making the search process stable and robust. (2) We propose sliding window ranking to estimate the optimal action efficiently. (3) We propose state retrieval to re-use the feedback sentences from the search history.}
Our experimental results demonstrate the performance of our model on three datasets. In summary, DQrank provides a crucial step toward applying RL algorithms in search tasks while opening up promising directions for future work.

\section{Limitations}
DQrank shows strengths in extracting interaction information better than the previous approaches. However, sentence-level feedback is still hard to obtain in most online search scenarios. Two solutions may overcome this limitation. First, joining more attention-detecting mechanisms into the search engines can help obtain sentence-level feedback efficiently, such as mouse tracking and eye tracking. Second, mixing with document-level feedback can help update the model when sentence-level feedback is absent. 
%\section{Ethical use of data and informed consent}
% All the data involved human participants comply with the ACM Code of Ethics and Professional Conduct and international and national standards. The volunteers are anonymous and informed of the usage of the data. 

\bibliographystyle{ACM-Reference-Format}
\bibliography{main-sigconf}

\newpage

\end{document}